\def\paperlanguage{}
\pdfoutput=1 

\documentclass[letterpaper, 10 pt, conference]{ieeeconf}  

\usepackage{bm}
\usepackage{cite}
\usepackage{flushend}
\include{preamble}

\newcommand{\ctext}[1]{\raise0.2ex\hbox{\textcircled{\scriptsize{#1}}}}

\IEEEoverridecommandlockouts                              

\title{\LARGE \textbf
  {
    \switchlanguage%
    {%
      MEVIUS: A Quadruped Robot Easily Constructed through E-Commerce with Sheet Metal Welding and Machining
    }%
    {%
      MEVIUS: 板金溶接と切削によりE-Commerceで\\簡単に構築可能な4脚ロボット
    }%
  }
}

\author{Kento Kawaharazuka$^{1}$, Shintaro Inoue$^{1}$, Temma Suzuki$^{1}$, Sota Yuzaki$^{1}$, Shogo Sawaguchi$^{1}$,\\Kei Okada$^{1}$, and Masayuki Inaba$^{1}$
  \thanks{$^{1}$ The authors are with the Department of Mechano-Informatics, Graduate School of Information Science and Technology, The University of Tokyo, 7-3-1 Hongo, Bunkyo-ku, Tokyo, 113-8656, Japan.
    {\texttt\small [kawaharazuka, s-inoue, t-suzuki, yuzaki, sawaguchi, k-okada, inaba]@jsk.t.u-tokyo.ac.jp}
  }
}
\begin{document}

\maketitle
\thispagestyle{empty}
\pagestyle{empty}

\begin{abstract}
  \switchlanguage%
  {%
    Quadruped robots that individual researchers can build by themselves are crucial for expanding the scope of research due to their high scalability and customizability.
    These robots must be easily ordered and assembled through e-commerce or DIY methods, have a low number of components for easy maintenance, and possess durability to withstand experiments in diverse environments.
    Various quadruped robots have been developed so far, but most robots that can be built by research institutions are relatively small and made of plastic using 3D printers.
    These robots cannot withstand experiments in external environments such as mountain trails or rubble, and they will easily break with intense movements.
    Although there is the advantage of being able to print parts by yourself, the large number of components makes replacing broken parts and maintenance very cumbersome.
    Therefore, in this study, we develop a metal quadruped robot MEVIUS, that can be constructed and assembled using only materials ordered through e-commerce.
    We have considered the minimum set of components required for a quadruped robot, employing metal machining, sheet metal welding, and off-the-shelf components only.
    Also, we have achieved a simple circuit and software configuration.
    Considering the communication delay due to its simple configuration, we experimentally demonstrate that MEVIUS, utilizing reinforcement learning and Sim2Real, can traverse diverse rough terrains and withstand outside experiments.
    All hardware and software components can be obtained from \href{https://github.com/haraduka/mevius}{\textcolor{magenta}{github.com/haraduka/mevius}}.
  }%
  {%
    研究者個人が自身の手で製作可能な4脚ロボットは, その拡張性やカスタマイズ性が高く, 研究の幅を広げるために重要である.
    それらのロボットには, 自主制作やE-Commerceにより簡易に発注・組み立てできること, 部品点数が少なくメンテナンスが容易なこと, 多様な環境での実験に耐えうる耐久性が重要である.
    これまで様々な4脚ロボットが開発されてきたが, 特に研究機関で自身の手で製作可能なロボットは, そのほとんどが3Dプリンタによるプラスチック製であり, 比較的小さい.
    これらのロボットはプラスチックゆえに, コンクリートの床や瓦礫のような不整地などの外部環境の実験に耐えられるものではなく, 激しい動作をすればすぐに破損してしまう.
    また, 自身で部品をプリントできるメリットはあるものの, 部品点数も多く破損した部品の交換やメンテナンスを考慮すると非常に手間がかかる.
    そこで本研究では, 4脚ロボットに必要な部品の最小セットを考慮しながら, ほとんどの部品が金属であり, かつe-commerceを通して構築可能なロボットMEVIUSを開発する.
    金属切削と板金溶接, その他off-the-shelfな部品のみで構築可能かつシンプルな回路構成・ソフトウェア構成を目指した.
    シンプルな構成ゆえの通信遅延を考慮した強化学習とSim2Realにより, 多様な不整地を踏破し, 外部環境での実験に耐え得ることを実験的に示した.
    個人や研究室単位で一から簡単に作成でき, 実環境にも耐えられる4脚ロボットが開発されることにより, 研究者によるカスタマイズや新しい機構の追加に基づき, さらなる面白い研究が盛んに行われることを期待する.
    本研究の全てのハードウェアとソフトウェアは\href{https://github.com/haraduka/mevius}{\textcolor{magenta}{github.com/haraduka/mevius}}から得ることができる.
  }%
\end{abstract}

\begin{figure}[t]
  \centering
  \includegraphics[width=0.95\columnwidth]{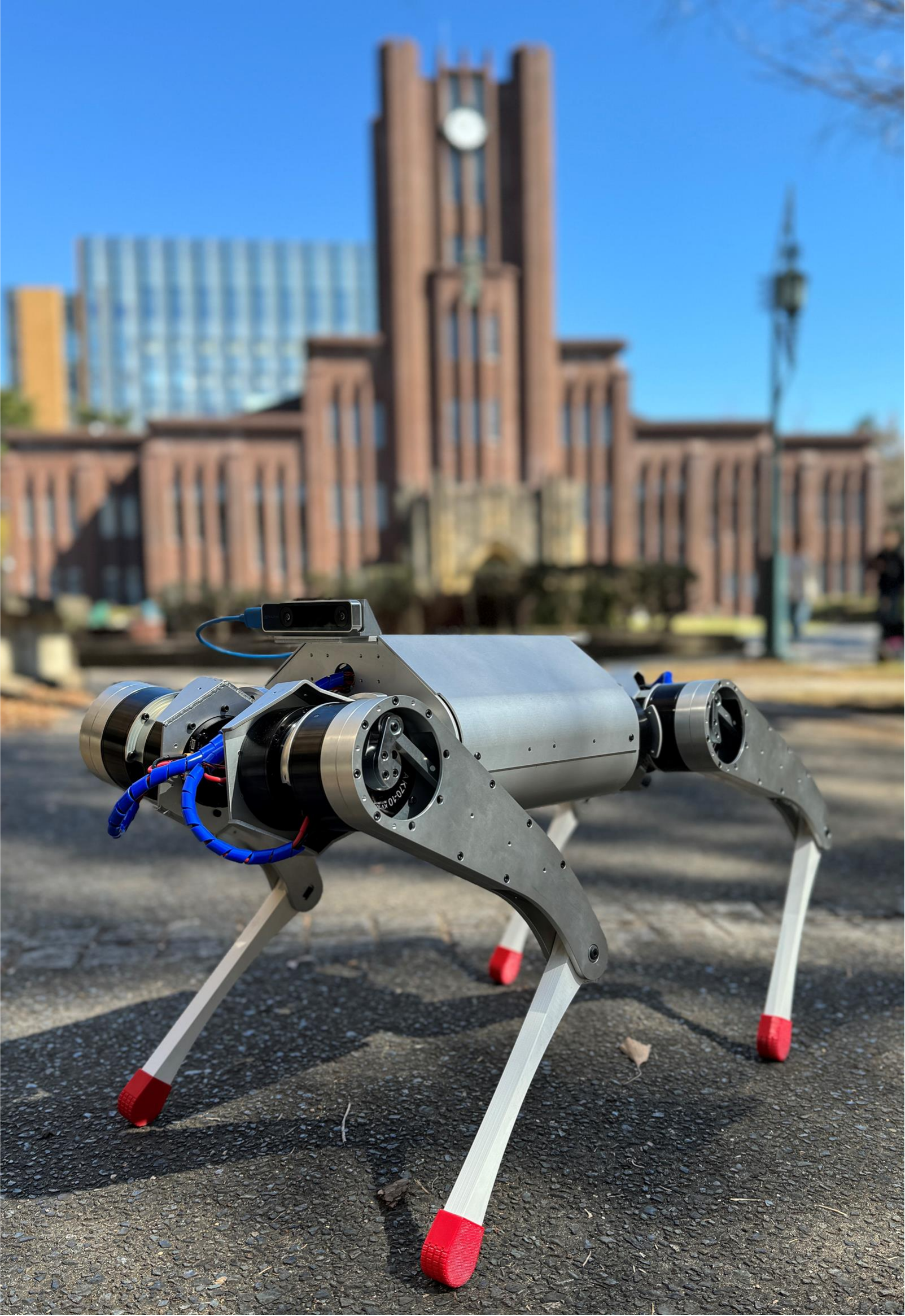}
  \vspace{-1.5ex}
  \caption{Overview of MEVIUS: a quadruped robot easily constructed through e-commerce with sheet metal welding and machining.}
  \label{figure:mevius}
  \vspace{-3.0ex}
\end{figure}

\section{INTRODUCTION}\label{sec:introduction}
\switchlanguage%
{%
  Various quadruped robots have been developed so far \cite{fujita1998quadruped, katz2019minicheetah, hutter2016anymal}, and advancements in model predictive control \cite{grandia2023nmpc} and reinforcement learning \cite{hwangbo2019anymal} have significantly contributed to their progress.
  Particularly, there is active development of robots capable of withstanding real-world usage, including walking on large steps and diverse uneven terrains \cite{lee2020anymal, miki2022anymal}.
  Consequently, a variety of quadruped robots, such as ANYMAL \cite{hutter2016anymal}, Unitree Go1 \cite{unitree2022go1}, and DEEPRobotics Lite3 \cite{deeprobotics2023lite3}, are now available for purchase worldwide.
  However, these robots have limitations in terms of accessibility to lower layers and customization, restricting the breadth of research.
  If a quadruped robot that can be easily constructed from scratch is available, it would significantly broaden the scope of research.

  For a robot that can be developed by individual researchers, three aspects are important: (i) it should be easy to order and assemble through e-commerce or DIY methods, (ii) have a small number of parts to facilitate easy maintenance, and (iii) possess durability to withstand experiments in various environments.
  Currently, representative examples of quadruped robots that can be easily created at the individual or research lab level include Solo \cite{grimminger2020solo, leziart2021solo12} and PAWDQ \cite{joonyoung2021pawdq}.
  The use of 3D printers and off-the-shelf motors and motor drivers have enabled the simplified creation of various small quadruped robots.
  However, when it comes to larger quadruped robots consisting of metal components, it becomes troublesome for individuals to build, requiring multiple steps of communication with various manufacturers.
  Especially for components like the torso and certain leg parts, which can be large or complex, machining might be difficult or financially challenging.
  Therefore, most open-source quadruped robots are small plastic robots created with 3D printing \cite{grimminger2020solo, leziart2021solo12, joonyoung2021pawdq, kau2021stanfordpupper, kau2019stanforddoggo, garc2020charlotte, sprowtz2018oncilla, lohmann2012aracna, rahme2021quadruped}.
  While these robots are easy to create, they cannot withstand experiments in external environments such as mountain trails or rubble, and they will easily break with intense movements.
  As the size increases, the number of parts also increases, making the replacement of broken parts and maintenance very labor-intensive.
  Consequently, we believe that if researchers could individually create larger quadruped robots made of metal with a small number of parts, it would facilitate design optimization of the robots and the addition of new mechanisms, fostering further interesting research opportunities for real-world use.
}%
{%
  これまで様々な4脚ロボットが開発されてきており\cite{fujita1998quadruped, katz2019minicheetah, hutter2016anymal}, モデル予測制御\cite{grandia2023nmpc}や強化学習\cite{hwangbo2019anymal}の進歩から, その発展は著しい.
  特に, 大きな段差や多様な不整地での歩行を含め, 実環境での利用に耐えうるロボットの開発が活発に行われている\cite{lee2020anymal, miki2022anymal}.
  これに伴い, 現在ではANYMAL \cite{hutter2016anymal}やUnitree Go1 \cite{unitree2022go1}, DEEPRobotics Lite3 \cite{deeprobotics2023lite3}など, 多様な4脚ロボットが世界で購入可能となってきた.
  しかし, 低レイヤを触りにくい, 自分で身体をカスタマイズしにくいなど, これらのロボットでは研究の幅は限られている.
  もし個人や研究室単位で一から簡単に作成できる4脚ロボットがあれば, 研究の幅は大きく広がるであろう.

  研究者個人で開発可能なロボットには, 自主制作やE-Commerceにより簡易に発注・組み立てできること, 部品点数が少なくメンテナンスが容易なこと, 多様な環境での実験に耐えうる耐久性の3つが重要である.
  現在, 個人や研究室単位で簡単に作成できる4脚ロボットとしては, Solo \cite{grimminger2020solo, leziart2021solo12}やPAWDQ \cite{joonyoung2021pawdq}が挙げられる.
  3Dプリンタとoff-the-shelfのモータ・モータドライバにより, 簡易に様々な小型の4脚ロボットが作られるようになってきた.
  しかし, 金属部品からなる一回り大きなサイズの4脚ロボットとなると, 個人で作るのは難しく, 多数の切削部品を見積もりに出す必要がある.
  特に胴体部品や脚の一部の部品は大きかったり複雑であったりするため, 切削が難しい場合や予算的に厳しい場合がある.
  そのため, オープンソースで開発可能なほとんどの4脚ロボットは基本構造に3Dプリンタを用いたプラスチック製の小型ロボットである\cite{grimminger2020solo, leziart2021solo12, joonyoung2021pawdq, kau2021stanfordpupper, kau2019stanforddoggo, garc2020charlotte, sprowtz2018oncilla, lohmann2012aracna, rahme2021quadruped}.
  これらのロボットは作成が容易な一方で, 実環境での利用には耐久性が足りない.
  プラスチックゆえに, コンクリートの床や瓦礫のような不整地などの外部環境の実験に耐えられるものではなく, 激しい動作をすればすぐに破損してしまう.
  サイズを大きくしていくと部品点数も増え, 破損した部品の交換やメンテナンスを考慮すると非常に手間がかかる.
  そこで, 金属で構成されるようなサイズの大きな4脚ロボットを少数部品かつ研究者個人で容易に作成可能にすることができれば, それらの設計最適化や新しい機構の追加等が容易であり, さらなる面白い研究が盛んに行われるであろうと考えた.
}%

\switchlanguage%
{%
  In this study, we developed a metal quadruped robot MEVIUS by minimizing the number of components from the body to the legs to only 10 types (excluding mirror parts), as shown in \figref{figure:mevius}.
  While the configuration of this robot is not significantly different from a typical quadruped robot, it is constructed easily with off-the-shelf, low gear ratio, and high torque servo motors and a minimal number of components that can be ordered through e-commerce.
  With the sole exception of the 3D printed part for the calf link, all other components can be obtained from the e-commerce site and the service of machining and sheet metal processing, ``meviy'' \cite{misumi2024meviy}.
  The calf link is given high strength using POTICON (Potassium Titanate Compound) NTL34M.
  In particular, the use of sheet metal welding allows for the cost-effective construction of large body parts and complex leg components as a single unit, leading to a significant reduction in the number of components.
  Additionally, the circuit configuration is very simple, eliminating the need to design or order custom circuits.
  It only uses a single-channel CAN-USB interface, which may cause some communication delays during control.
  However, by employing reinforcement learning that takes communication delays into account, we have successfully achieved walking movements in various real-world environments.
  All hardware and software components can be obtained from \href{https://github.com/haraduka/mevius}{\textcolor{magenta}{github.com/haraduka/mevius}}.
}%
{%
  本研究では, 極限まで部品数を減らし, ミラー部品を除くと10種類の部品のみで胴体から脚までが構成可能な金属製の4脚ロボットMEVIUSを開発した(\figref{figure:mevius}).
  このロボットの構成は一般的な4脚ロボットと特に大きな変わりはないものの, off-the-shelfな低減速比高トルクのサーボモータとe-commerceで発注可能な少数の部品のみで簡易に構成される.
  脚の3Dプリント部品以外については, 特にe-commerceサイトであるMISUMIとその切削・板金加工サービスであるmeviy \cite{misumi2024meviy}のみを用いて全ての部品を揃えることができるため, 非常にお手軽な設計である.
  脚についても, POTICON (Potassium Titanate Compound) NTL34Mを用いた高強度な構成となっている.
  特に, 板金溶接を用いることで, 巨大な胴体部品や複雑な脚部品を一部品として構築することができ, 大きく部品点数を削減できている.
  また, 非常にシンプルな回路構成をしており, 特に回路を設計したり発注したりする手間もない.
  1チャネルのCAN-USBインターフェースしか用いておらず, 制御の際には通信遅延が多少問題となるが, 通信遅延を考慮した強化学習を行い, 多様な実環境における歩行動作実現に成功している.
  個人や研究室単位で一から簡単に作成でき, 実環境にも耐えられる4脚ロボットが開発されることにより, 研究者によるカスタマイズや新しい機構の追加に基づき, さらなる面白い研究が盛んに行われることを期待する.
  本研究の全てのハードウェアとソフトウェアは\href{https://github.com/haraduka/mevius}{\textcolor{magenta}{github.com/haraduka/mevius}}から得ることができる.
}%

\section{Design and Configuration of MEVIUS} \label{sec:musculoskeletal-humanoids}
\begin{figure*}[t]
  \centering
  \includegraphics[width=1.95\columnwidth]{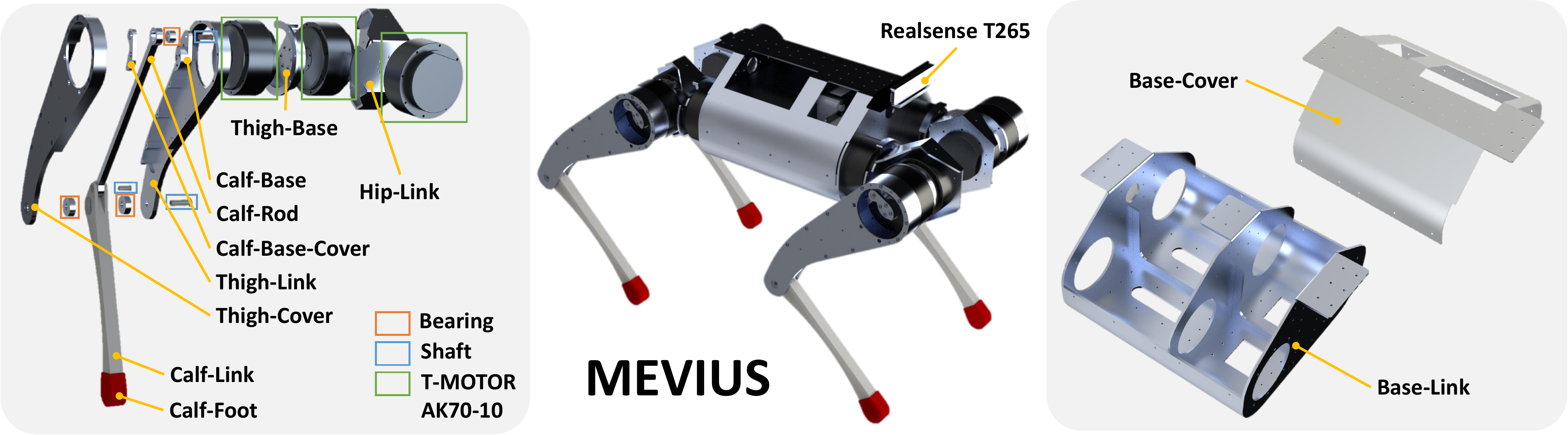}
  \vspace{-1.0ex}
  \caption{The design overview of MEVIUS. The basic structure is composed of a total of 10 types of components, excluding mirror parts and feet.}
  \label{figure:design}
  \vspace{-0.5ex}
\end{figure*}

\switchlanguage%
{%
  First, we consider the structure of a general quadruped robot and its minimal configuration.
  Next, we will elaborate on the detailed design of MEVIUS and compare it to existing quadruped robots.
  Finally, we will discuss the circuitry and the configuration of the learning-based software for MEVIUS.
  The overall design is illustrated in \figref{figure:design}.
}%
{%
  始めに, 一般的な4脚ロボットの構造と, その最小構成について考える.
  次に, MEVIUSの詳細な設計について述べ, これを既存の4脚ロボットと比較する.
  最後に, MEVIUSの回路と学習型ソフトウェア構成について述べる.
  なお, 設計の全体像は\figref{figure:design}に示す.
}%

\begin{figure}[t]
  \centering
  \includegraphics[width=0.95\columnwidth]{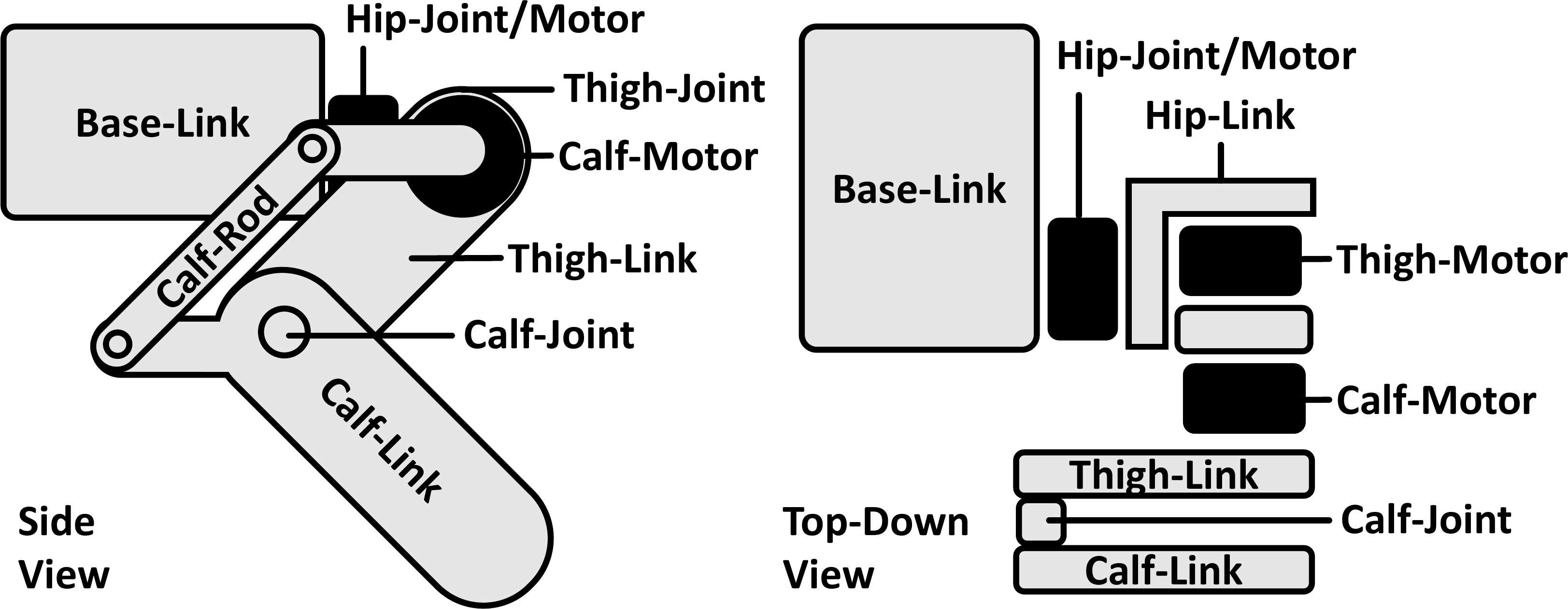}
  \vspace{-1.0ex}
  \caption{The basic structure of quadruped robots with a parallel link mechanism.}
  \label{figure:basic-structure}
  \vspace{-3.0ex}
\end{figure}

\subsection{The Basic Structure of Quadruped Robots and Minimum Configuration} \label{subsec:basic-structure}
\switchlanguage%
{%
  Quadruped robots typically have a configuration with three motors on each leg, totaling 12 motors.
  As illustrated in \figref{figure:basic-structure}, the essential links include the central Base-Link, Hip-Link extending from Base-Link, and Thigh-Link and Calf-Link of the leg.
  The joints between these links are referred to as Hip-Joint, Thigh-Joint, and Calf-Joint.
  The motors driving these joints are also named Hip-Motor, Thigh-Motor, and Calf-Motor.
  Assuming a conventional arrangement, these joints are aligned in the order of roll, pitch, and pitch.
  While structures utilizing belts, such as Solo \cite{grimminger2020solo, leziart2021solo12}, can be considered, for robots larger than Mini Cheetah \cite{katz2019minicheetah}, belts may not withstand the joint torque.
  In such cases, a parallel link structure is often adopted, similar to Unitree Go1 \cite{unitree2022go1}.

  Here, we discuss the minimum number of components required, especially when using parallel links.
  Note that only structural components related to a single leg from the body are considered, excluding mirror components, and we assume the use of servo motors with integrated circuitry and do not consider off-the-shelf items such as rotating shaft, bearings, washers, etc.
  Also, we disregard the potential cost implications resulting from overly large or complex component structures.
  First, the minimum number of components of Base-Link is two.
  One serves as the foundation, while the other plays a covering role, such as enclosing the circuit.
  While it is possible to construct the entire base with a single component, practical considerations (including maintenance and circuit protection) make this unrealistic.
  Next, Hip-Link, connecting Hip-Motor and Thigh-Motor, requires only one component in the minimal configuration.
  Subsequently, one component is necessary to connect Thigh-Motor and Calf-Motor.
  Finally, in addition to Thigh-Link, Calf-Link, and the rod of the parallel link (Calf-Rod), one component is required to connect Calf-Motor and Calf-Rod.

  Here, we consider the cost of the components.
  First, the connection between Thigh-Link and Calf-Link, Calf-Link and Calf-Rod, and Calf-Rod and Calf-Motor requires bearings.
  Particularly, for the connection between Thigh-Link and Calf-Link, where a significant load is applied, a double-bearing arrangement is desirable.
  In order to reduce cost, when constructing Thigh-Link through metal machining, it is advisable to divide this Thigh-Link into two parts, creating a structure that holds the Calf-Link from both sides.
  Regarding the connection components for Calf-Rod and Calf-Motor, it is desirable to split the components into two parts, considering the option of grasping the Calf-Rod from both sides to achieve a dual-hold configuration.
  Moreover, Base-Link is significantly large, and Hip-Link tends to become complex due to connecting motors in different directions, often requiring component segmentation.
  In summary, the minimum configuration necessitates 8 types of components, and considering cost implications, approximately 10-12 types of components are required.
}%
{%
  まず4脚ロボットの一般的な構造について述べる
  これらは各脚に3つのモータ, 計12個のモータを有する構成が一般的である.
  \figref{figure:basic-structure}に示すように, 主に必要なリンクは中央のBase-Link, Base-Linkから伸びる脚のHip-Link, Thigh-Link, Calf-Linkであり, それらの間の関節をHip-Joint, Thigh-Joint, Calf-Joint それらの関節を駆動するMotorをHip-Motor, Thigh-Motor, Calf-Motorと呼ぶこととする.
  これらの関節はロール, ピッチ, ピッチの順に, 一般的な並びをしていると仮定する.
  Solo \cite{grimminger2020solo, leziart2021solo12}のようなベルトを使った構造も考えられるが, Mini Cheetah \cite{\cite{katz2019minicheetah}}程度よりもサイズが大きくなると, ベルトではトルクに耐えられなくなるため, Unitree Go1 \cite{unitree2022go1}と同様な平行リンク構造を採用する場合が多くなる.

  ここでは, 特に平行リンクを使った場合について, どれだけの部品点数が必要か, その最小構成について述べる.
  なお, 胴体から一つの脚に関する構造部品だけを考え, ミラー部品は考慮しないこと, 回路構成が一体となったサーボモータを使用すること, 回転軸やベアリング, ワッシャなどの既成品については考えないことを前提とする.
  また, その際に部品の構造が大き過ぎたり, 複雑過ぎたりすることによるコストは考えない.
  まず, Base-Linkは2部品まで減らすことができる.
  これは, 一つが基礎となり, もう一つは回路を覆うなどのカバーの役割をする必要があるためである.
  Base-Link一つで構成することもできるが, メンテナンスや回路の保護を考えると現実的ではない.
  次に, Hip-LinkはHip-MotorとThigh-Motorをつなげる部品であるため, 一つの部品が最小構成である.
  次に, Thigh-MotorとCalf-Motorをつなげるのに一部品が必要である.
  最後に, Thigh-Link, Calf-Link, 平行リンクのロッド(Calf-Rod)に加え, Calf-MotorとCalf-Rodを接続する一部品が必要である.

  ここでコストについて考える.
  まず, Thigh-LinkとCalf-Linkの接続, Calf-LinkとCalf-Rodの接続, Calf-RodとCalf-Motorの接続には, ベアリングを要する.
  特に, Thigh-LinkとCalf-Linkの接続には大きな負荷がかかるためベアリングは両持ちが望ましい.
  金属切削でThigh-Linkを構成する場合, コストを下げるためにも, このThigh-Linkは2つに分け, Calf-Linkを両側から押さえる構造が望ましい.
  Calf-RodとCalf-Motorの接続部品については, Calf-Rodを両側から押さえて両持ちにすることを考えると, コスト次第で部品を2つに分けることが望ましい.
  また, Base-Linkは非常に大きく, Hip-Linkは異なる方向のモータ同士を接続するため複雑になりやすく, 部品を分割する必要がある場合は多い.
  よってこれらをまとめると, 最小構成で8種類, コストを考えると10-12種類程度の部品が必要である.
}%

\begin{figure}[t]
  \centering
  \includegraphics[width=0.7\columnwidth]{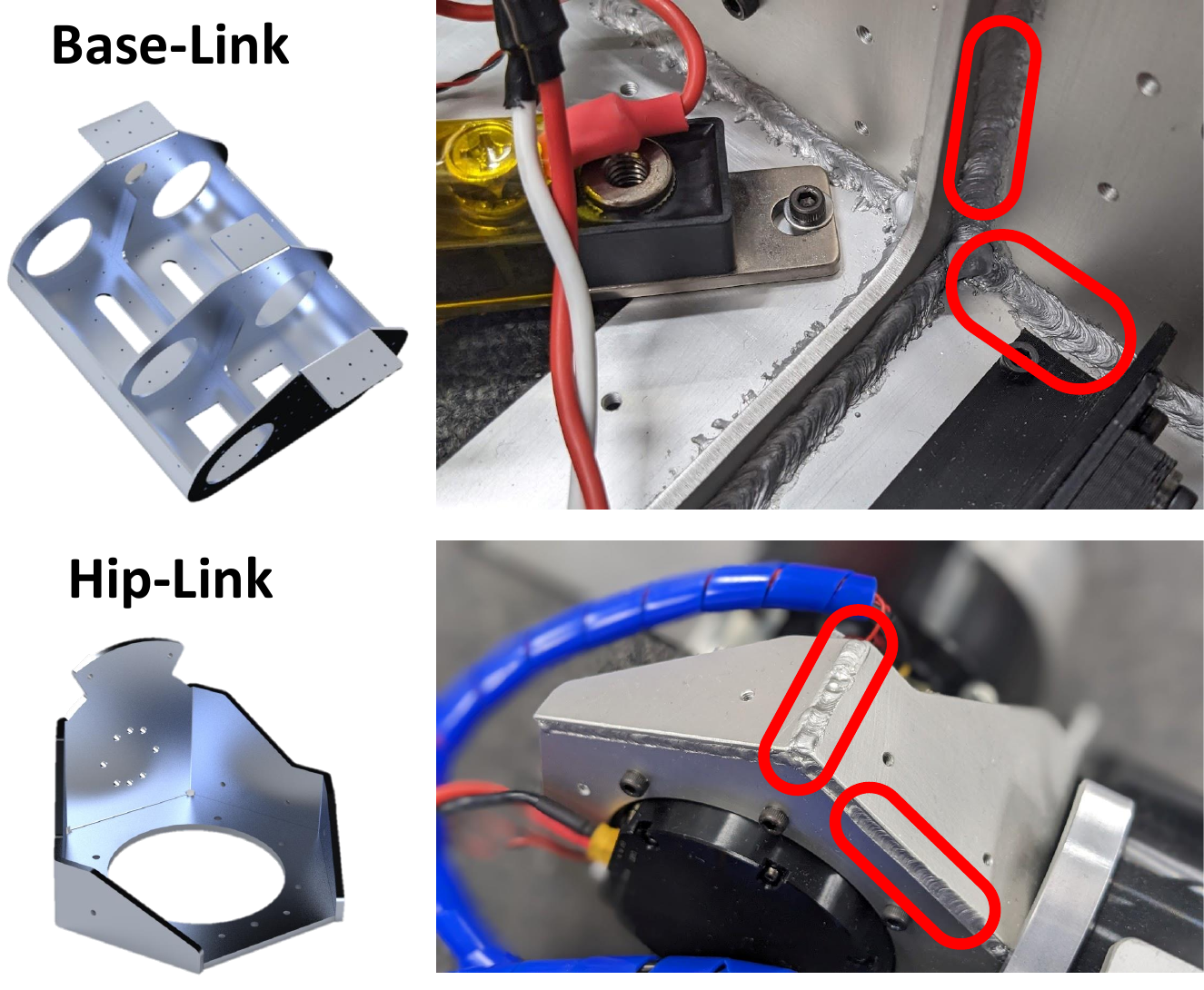}
  \vspace{-1.0ex}
  \caption{Base-Link and Hip-Link constructed through sheet metal welding.}
  \label{figure:welding}
  \vspace{-3.0ex}
\end{figure}

\begin{table*}[t]
  \centering
  \caption{Comparison between existing quadruped robots and MEVIUS}
  \begin{tabular}{l|ccccccc}
    Name & Weight & Leg Length & Materials & Open/Closed (CAD) & Off-the-shelf (Circuit) & Maximum Torque \\ \hline
    Mini Cheetah \cite{katz2019minicheetah}    & 9.0 kg  & 0.20 m & Metal                  & Closed        & No           & 17 Nm  \\
    ANYMAL \cite{hutter2016anymal}             & 30.0 kg & 0.25 m & Metal                  & Closed        & No           & 40 Nm  \\
    Solo-12 \cite{leziart2021solo12}           & 2.5 kg  & 0.16 m & Plastic                & \textbf{Open} & No           & 2.5 Nm \\
    Stanford Doggo \cite{kau2019stanforddoggo} & 4.8 kg  & 0.16 m & CFRP / Metal / Plastic & \textbf{Open} & \textbf{Yes} & 4.8 Nm \\
    PAWDQ \cite{joonyoung2021pawdq}            & 12.7 kg & 0.22 m & Plastic                & \textbf{Open} & \textbf{Yes} & 21 Nm  \\
    MEVIUS (This Study)                        & 15.5 kg & 0.25 m & Metal / POTICON        & \textbf{Open} & \textbf{Yes} & 25 Nm  \\
  \end{tabular}
  \label{table:comparison}
\end{table*}

\subsection{Design of MEVIUS} \label{subsec:mevius-design}
\switchlanguage%
{%
  Based on \secref{subsec:basic-structure}, we elaborate on the detailed design of MEVIUS.
  12 servo motors AK70-10 (T-Motor) with a gear ratio of 10:1 are installed, three on each leg.
  All components, except for Calf-Link, are made of aluminum, and Calf-Link is a 3D-printed part using the high-strength nylon-based plastic, POTICON (Potassium Titanate Compound) NTL34M.
  The link structure of MEVIUS is composed of a total of 10 types of parts, dividing two of the parts in the minimum configuration in \secref{subsec:basic-structure} considering cost (excluding bearings, rotating shafts, sensors, and Calf-Foot covering Calf-Link).
  All motors are interconnected through the Base-Link, the Hip-Link connecting Hip-Motor and Thigh-Motor, and the Thigh-Base connecting Thigh-Motor and Calf-Motor.
  For cost efficiency, Thigh-Link and Thigh-Cover are split, as well as Calf-Base (connecting Calf-Rod and Calf-Motor) and Calf-Base-Cover.
  The basic leg structure is constructed using Calf-Rod, Calf-Link, Thigh-Link, Thigh-Cover, Calf-Base, and Calf-Base-Cover.
  Calf-Link, being subject to frequent changes in shape, length, and tip design for research purposes, is constructed as a single part without splitting using POTICON filament via 3D printing.
  Thus, a total of 10 types of parts are required for the basic structure.
  Additionally, two ball bearings are inserted into Thigh-Link, two needle bearings are inserted into Calf-Rod, and two rotating shafts are needed.
  To prevent slipping, Calf-Foot made through 3D printing of TPU (Thermoplastic Polyurethane) is applied as an end effector, and Intel Realsense T265 is mounted on the head to obtain linear and angular velocity of Base-Link.

  An important aspect is the fabrication of large or complex Base-Link and Hip-Link as a single part using sheet metal welding, as illustrated in \figref{figure:welding}.
  Sheet metal welding enables the economical and high-strength manufacturing of large and complex parts.
  In addition to machining and sheet metal processing, meviy \cite{misumi2024meviy}, an e-commerce platform by MISUMI, allows for automated quotation of sheet metal welding parts directly from CAD files.
  This eliminates the need for creating drawings and getting quotations, and enables the ordering of the metal parts solely through CAD files.
  When creating a structure similar to Hip-Link by machining with meviy, the cost is approximately \$400, whereas utilizing sheet metal welding allows for a significant cost reduction, totaling around \$100.
  Note that all other mechanical components can be purchased as off-the-shelf products through e-commerce services.
}%
{%
  \secref{subsec:basic-structure}を踏まえ, MEVIUSの詳細な設計について述べる.
  一つの脚で3つずつ, 計12個の減速比10:1なサーボモータAK70-10 (T-Motor)を搭載する.
  Calf-Link以外は全てアルミニウムにより構成されており, Calf-Linkは高強度なナイロンベースのプラスチックであるPOTICON (Potassium Titanate Compound) NTL34Mを利用した3Dプリンタ造形部品となっている.
  MEVIUSのリンク構造は, \secref{subsec:basic-structure}の最小構成にコストを考慮し2つの部品を分割した, 計10種類の部品により構成される(\secref{subsec:basic-structure}と同様にベアリングや軸, センサ, Calf-LinkをカバーするCalf-Footは除く).
  まず, Base-Link, Hip-MotorとThigh-Motorを繋ぐHip-Link, Thigh-MotorとCalf-Motorを繋ぐThigh-Baseにより全てのモータが接続される.
  次に, コストのために分割したThigh-Linkとそれを覆うThigh-Cover, 同様にコストのために分割したCalf-Base(Calf-RodとCalf-Motorを接続する部品)とCalf-Base-Cover, そしてCalf-RodとCalf-Linkにより基本の脚構造が構築される.
  Calf-Linkは研究に応じてその形や長さ, 先端の形状などを頻繁に変更する可能性があるため, POTICONフィラメントの3Dプリンタ造形を用いることで, 分割することなく一つの部品として構成している.
  よって, 基本の構造に必要な部品は合計で10種類となっている.
  これに加えて, 1脚につきThigh-Linkに挿入する玉軸受が2つ, Calf-Rodに挿入するニードルベアリングが2つ, 回転軸が2本必要である.
  また, 滑りを防止するため, TPUによる3Dプリンティングを用いたCalf-Footをエンドエフェクタにかぶせており, 頭部にはIntel Realsense T265が装着されている.

  ここで重要なことは, 大きくなりやすい, または複雑になりやすいBase-LinkとHip-Linkを\figref{figure:welding}に示すように, 板金溶接により一部品として構成している点である.
  板金溶接は予算を抑えつつ, 大きく複雑な部品を簡易かつ高強度に製造することが可能である.
  切削や板金加工はもちろんのこと, 現在ではMISUMIが展開するe-commerceであるmeviy \cite{misumi2024meviy}で, 自動でCADから板金溶接加工の部品見積もりを行うことができる.
  これにより, 人による見積もりや, やり取りを介さず, CADファイルのみで本研究の切削部品・板金部品を発注することが可能となっている.
  meviyでHip-Linkと同様な構造を切削で作る場合は約400ドルなのに対して, 板金溶接で作る場合は約100ドルと大きくコストを抑えることができる.
  なお, その他の機械部品は全てe-commerceサービスを通して購入することが可能な既成品である.
}%

\subsection{Comparison with Existing Quadruped Robots} \label{subsec:comparison}
\switchlanguage%
{%
  We present a comparison table (\tabref{table:comparison}) between MEVIUS developed in this research and other representative existing quadruped robots.
  The table includes information on overall weight, leg length, materials constituting the body and legs, availability of CAD files, whether circuits and motors are off-the-shelf or custom-made, and maximum torque of the motors.
  Notably, Mini Cheetah \cite{katz2019minicheetah} and ANYMAL \cite{hutter2016anymal} are highlighted as examples of robots that are not publicly disclosed.
  These structures are primarily composed of metal, demonstrating robust designs suitable for operation in diverse real-world environments.
  In contrast, Solo-12 \cite{leziart2021solo12}, Stanford Doggo \cite{kau2019stanforddoggo}, and PAWDQ \cite{joonyoung2021pawdq} are robots with publicly available CAD files.
  The structures of these robots are predominantly created using 3D printing with PLA or ABS resin.
  Stanford Doggo utilizes CFRP and metal, with 8 degrees of freedom and a configuration consisting only of pitch joints.
  Additionally, due to its complex structure combining belts and closed links, the number of components becomes quite extensive.
  Note that only Solo-12 requires the assembly of custom circuits.

  In comparison, MEVIUS is a robot with a larger body (15.5 kg and 0.25 m legs) than PAWDQ and Mini Cheetah, composed entirely of metal and POTICON (only Calf-Link).
  The maximum torque of MEVIUS is notably larger compared to other robots, excluding ANYMAL.
  All CAD parts are publicly available, and both circuits and motors are off-the-shelf components.
  Notably, the metal machining for each part is simplified, allowing for processing on only two surfaces, unlike the intricate designs of Mini Cheetah and ANYMAL.
}%
{%
  本研究で開発したMEVIUSと, その他の代表的な既存の4脚ロボットを比較したテーブルを\tabref{table:comparison}に示す.
  ここでは, 全体の重さ, 胴体と脚を成す基本構造の素材, 部品のCADが公開されているか否か, 回路やモータ等が既成品のみか否か, パワー電源電圧, モータの最大トルクを示している.
  まず, 一般公開されていないロボットとして, Mini Cheetah \cite{katz2019minicheetah}とANYMAL \cite{hutter2016anymal}が挙げられる.
  これらの構造は基本的に金属で構成されており, 外部の実環境でも利用できるような頑健な設計となっている.
  これに対して, CADファイルが公開されているロボットとして, Solo-12 \cite{leziart2021solo12}, Stanford Doggo \cite{kau2019stanforddoggo}, PAWDQ \cite{joonyoung2021pawdq}が挙げられる.
  これらのロボットの構造は基本的にPLAやABS樹脂の3Dプリンタによるものである.
  Stanford DoggoはCFRPと金属を利用しているが, 自由度が8でピッチ関節のみの構成であると同時に, ベルトと閉リンクを組み合わせた構造なため部品点数はかなり多くなってしまう.
  なお, Solo-12のみ回路は既成品ではなく, 回路図から部品を集め発注する必要がある.

  これらに対して, 本研究で開発するMEVIUSは15.5 kgとPAWDQやMini Cheetahよりも一回り大きな身体を持つロボットであり, かつ全身が金属とPOTICON(Calf-Linkのみ)で構成されている.
  それに伴い最大トルクもANYMALを除く他のロボットに比べると大きくなっている.
  また, 全てのCAD部品は公開されており, 回路やモータは全て既成品を利用したロボットとなっている.
  特に各金属切削部品については, 上下2面のみの加工が可能であり, これはMini CheetahやANYMALの複雑な設計とは異なる.
}%

\begin{figure}[t]
  \centering
  \includegraphics[width=0.95\columnwidth]{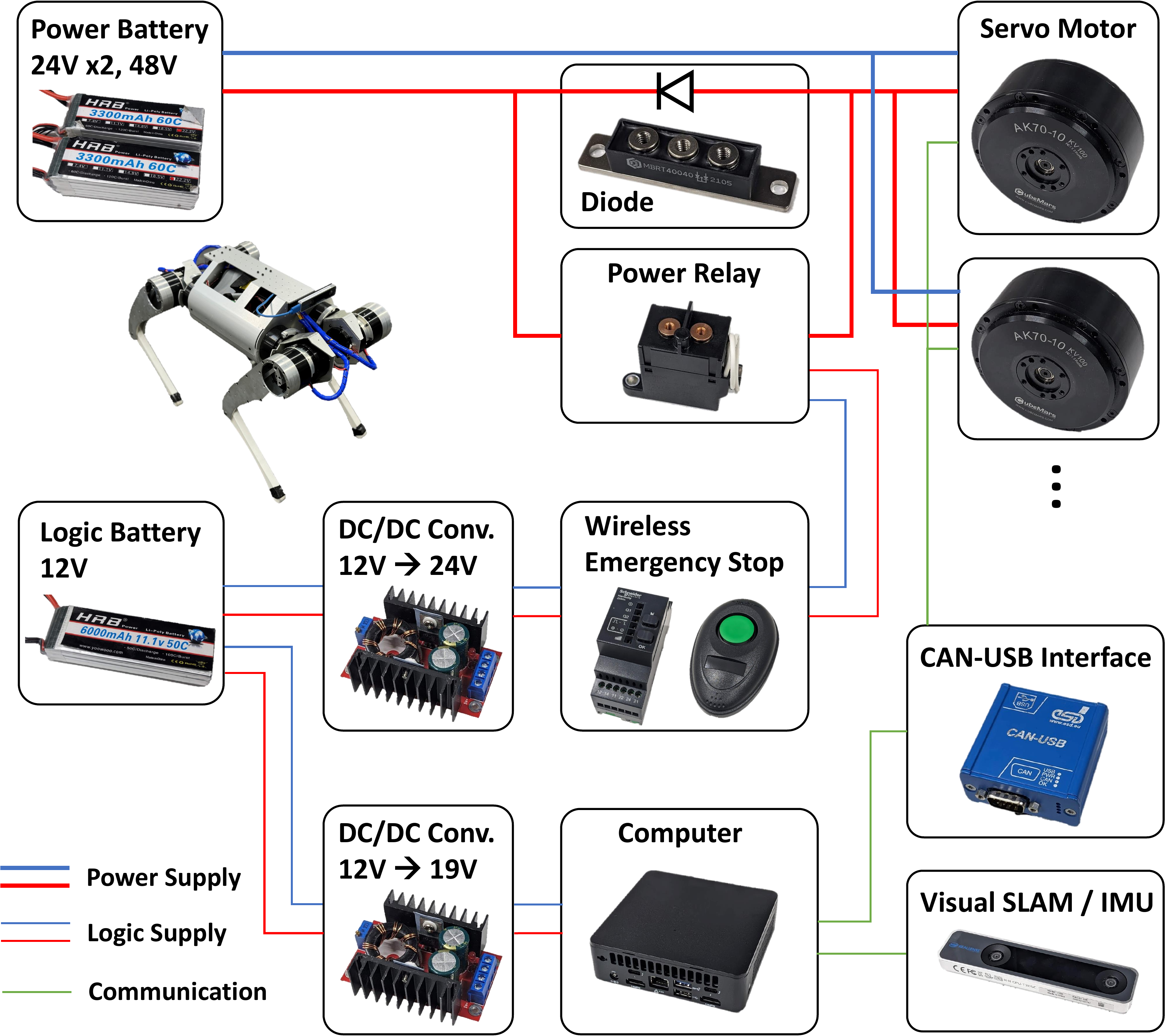}
  \vspace{-1.0ex}
  \caption{Circuit configuration of MEVIUS.}
  \label{figure:circuit}
\end{figure}

\begin{figure}[t]
  \centering
  \includegraphics[width=0.95\columnwidth]{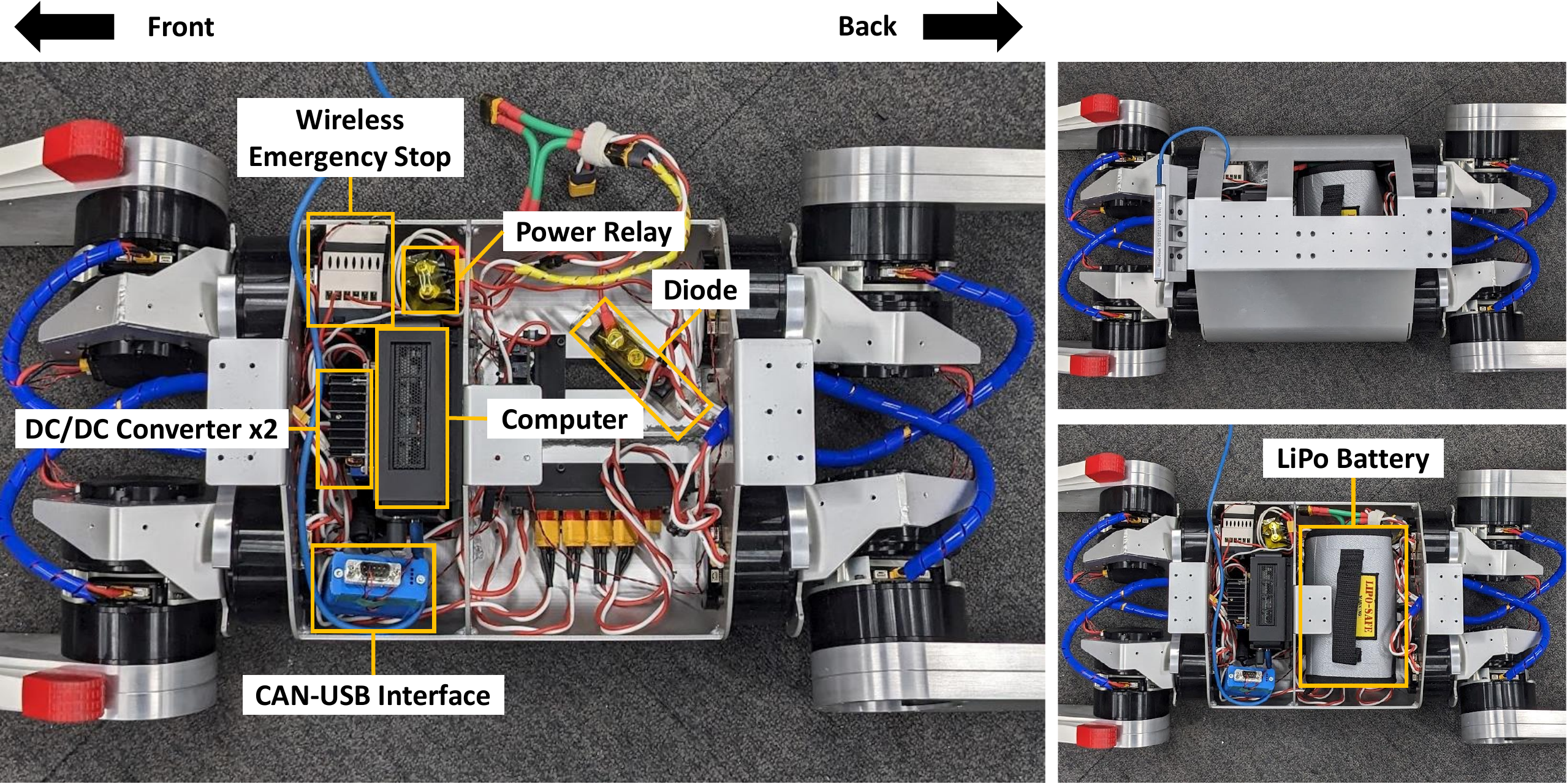}
  \vspace{-1.0ex}
  \caption{The arrangement of circuits in Base-Link of MEVIUS.}
  \label{figure:circuit2}
  \vspace{-3.0ex}
\end{figure}

\begin{table}[t]
  \centering
  \caption{Circuit-related Components for MEVIUS}
  \scalebox{0.9}{
    \begin{tabular}{l|cc}
      Device & Description & Quantity \\ \hline
      Servo Motor             & T-Motor AK70-10           & 12 \\
      Computer                & Intel NUC12WSKi7          & 1 \\
      CAN-USB Interface       & ESD CAN-USB/2             & 1 \\
      DC/DC Converter         & KOOKYE DC-DC Boost Module & 2 \\
      Wireless Emergency Stop & Harmony ZBRRA             & 1 \\
      Power Relay             & OMRON G9EN-1-UVD-DC24     & 1 \\
      Diode                   & GeneSiC MBRT40040         & 1 \\
      Visual SLAM / IMU       & Intel Realsense T265      & 1 \\
      Logic Battery           & HRB 3S 6000mAh 11.1V      & 1 \\
      Power Battery           & HRB 6S 3300mAh 22.2V      & 2
    \end{tabular}
    }
    \label{table:circuits}
\end{table}

\subsection{Circuit Configuration of MEVIUS}
\switchlanguage%
{%
  The circuit configuration of MEVIUS is illustrated in \figref{figure:circuit}.
  This circuit configuration can be considered the simplest setup utilizing CAN communication.
  The logic power is supplied by a 12V LiPo battery, while the motor power is supplied from two 24V LiPo batteries connected in series.
  The logic power is boosted to 24V and 19V, providing power to Wireless Emergency Stop and the computer, respectively.
  It is worth noting that the logic voltage could be set to 24V, reducing one DC/DC converter, but for future expandability, the logic voltage is set to the more common 12V.
  Regeneration from the servo motor to the battery is enabled by utilizing a diode, and Wireless Emergency Stop triggers the power relay.
  The servo motor used in this study, AK70-10 (T-Motor), has identical logic and power voltages, resulting in a configuration where the power to the entire logic system is cut in the event of an emergency stop.
  From the computer, each motor is connected through a daisy chain via a 1-channel CAN-USB interface.
  The internal arrangement of these circuit configurations within Base-Link is shown in \figref{figure:circuit2}.
  PC, Wireless Emergency Stop, CAN-USB interface, and others are consolidated at the front of the Base-Link, with LiPo batteries mounted at the rear.
  Finally, a list of commercially available components used in the circuit configuration is presented in \tabref{table:circuits}.
}%
{%
  MEVIUSの回路構成について\figref{figure:circuit}に示す.
  本回路構成は, CAN通信を使った最も簡単な構成と言えるだろう.
  電源については, ロジックを12VのLiPoバッテリーから, パワーを24VのLipoバッテリーを2つ直列に繋げたものから供給している.
  ロジック電源は24Vまたは19Vに昇圧され, それぞれ無線緊急停止とコンピュータに電源を供給する.
  なお, ロジック電圧も24VにすればDCDCコンバータを一つ減らすことができるが, 今後の拡張性のためロジック電圧は一般的な12Vに設定している.
  無線緊急停止は, ダイオードによってサーボモータからバッテリーへの回生を許しつつ, パワーリレーを駆動する.
  本研究で用いるサーボモータであるAK70-10 (T-Motor)はロジック電圧とパワー電圧が同一であるため, 緊急停止によってロジックごと電源が切れる構成となっている.
  コンピュータからは, 1チャンネルのCAN-USBインターフェースを通してデイジーチェーンで各モータに接続している.
  これら回路構成のBase-Link内部の配置を\figref{figure:circuit2}に示す.
  PCや緊急停止, CAN-USBインターフェースなどがBase-Linkの前方に集約され, 後方にはLiPoバッテリーが搭載される形となっている.
  最後に, \tabref{table:circuits}に回路構成で利用した既成品の一覧を示す.
}%

\begin{figure}[t]
  \centering
  \includegraphics[width=0.95\columnwidth]{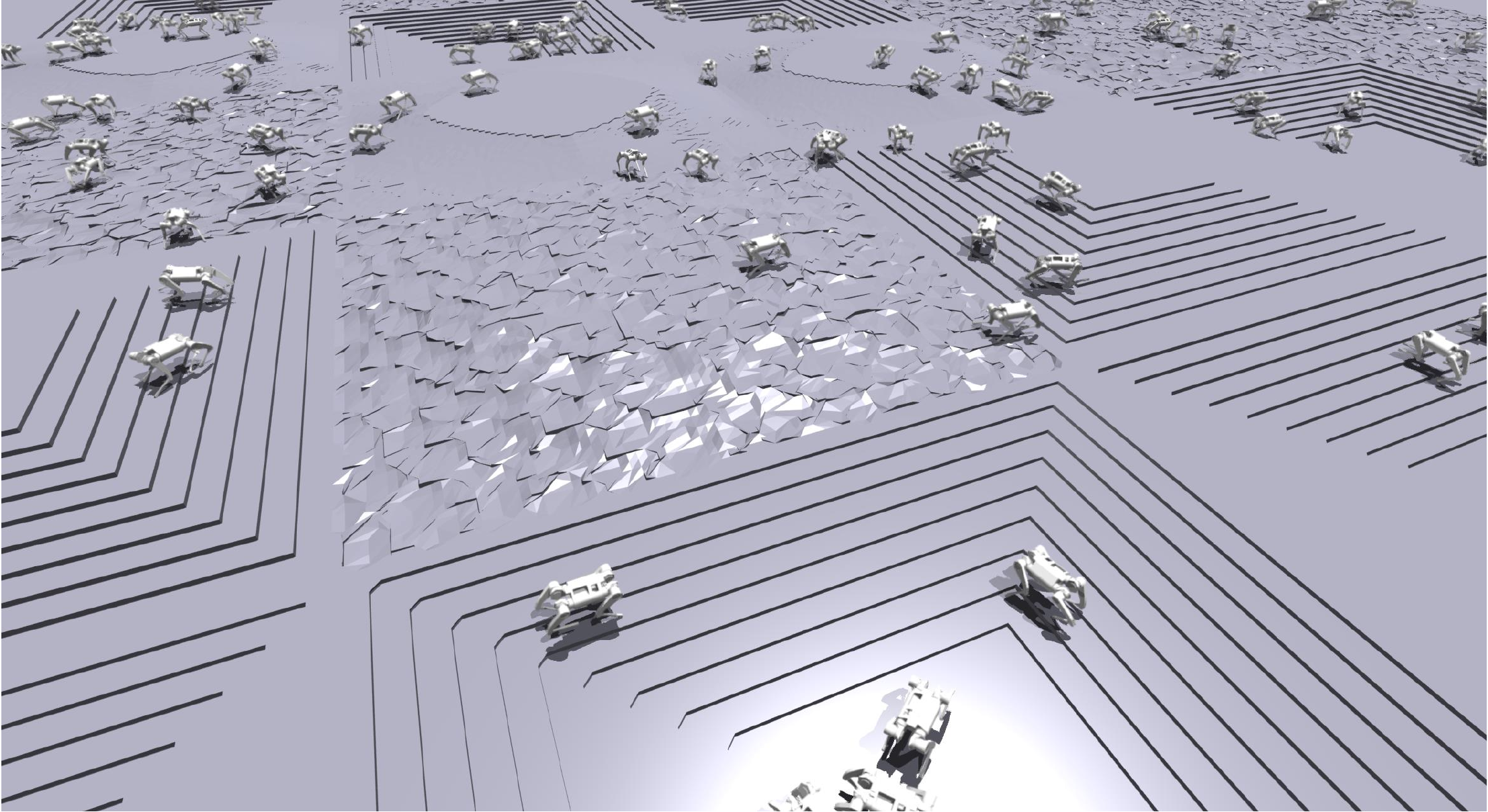}
  \vspace{-1.0ex}
  \caption{The simulation environment of IsaacGym for reinforcement learning of MEVIUS walking.}
  \label{figure:exp-sim}
  \vspace{-3.0ex}
\end{figure}

\begin{figure}[t]
  \centering
  \includegraphics[width=0.95\columnwidth]{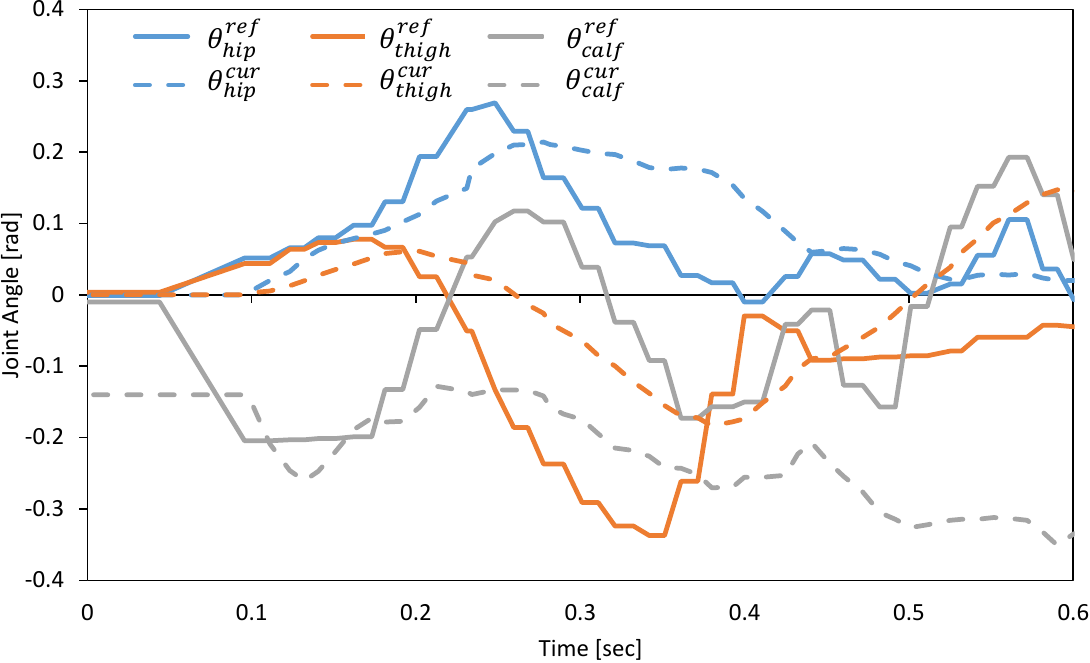}
  \vspace{-1.5ex}
  \caption{The tracking error of the front-right leg of MEVIUS when executing reinforcement learning without considering action delay.}
  \label{figure:exp-graph}
  \vspace{-3.0ex}
\end{figure}

\subsection{Control of MEVIUS}
\switchlanguage%
{%
  We will explain the control of MEVIUS.
  All programs are written in Python, and CAN communication is conducted through the Python Socket library.
  We introduced the model of MEVIUS into legged\_gym \cite{rudin2022leggedgym} for reinforcement learning, and the results are executed on the computer in MEVIUS.
  The learning environment is depicted in \figref{figure:exp-sim}.
  Here, to achieve the softest state while still being able to stand, the PD control gains ($K_p$, $K_d$) were set to (50, 2) for the Hip-Joint and Thigh-Joint, and (30, 0.2) for the Calf-Joint.
  For observations in RL, we followed the default settings of legged\_gym \cite{rudin2022leggedgym}; refer to \href{https://github.com/haraduka/mevius}{\textcolor{magenta}{github.com/haraduka/mevius}} for other parameters.

  A crucial aspect in reinforcement learning on MEVIUS is communication delay.
  To achieve simplicity in circuit configuration, we control 12 servo motors with a 1-channel CAN-USB interface.
  Consequently, the overall control cycle is 150Hz, and the execution cycle of reinforcement learning is 50Hz, running asynchronously as a thread.
  Compared to robots directly controlled by microcontrollers or FPGAs, there is a certain amount of communication delay.
  Executing reinforcement learning without considering communication delay and introducing it to the real robot would result in the robot's legs breaking due to severe vibrations.
  \figref{figure:exp-graph} illustrates the commanded joint angles $\bm{\theta}^{ref}_{\{hip, thigh, calf\}}$ and the actual joint angles $\bm{\theta}^{cur}_{\{hip, thigh, calf\}}$ of the right front leg during such a situation.
  In the simulator, the actual joint angles closely follow the commanded values, but on the real robot, a significant time-shift in the temporal direction is observed.
  To address this issue, we offset the reflection of actions in reinforcement learning by several steps.
  Preliminary experiments revealed that the communication delay in MEVIUS is approximately 0.02 to 0.06 seconds.
  Therefore, we conducted reinforcement learning by randomly delaying actions for 1 to 3 frames.
}%
{%
  最後にMEVIUSの制御について述べる.
  全てのプログラムはPythonによって記述され, CAN通信はPythonのSocketライブラリを通して行われる.
  legged\_gym\cite{rudin2022leggedgym}にMEVIUSのモデルを導入して強化学習を行い, その結果をMEVIUS上のコンピュータで実行している.
  その学習環境を\figref{figure:exp-sim}に示す.
  この際, ギリギリ立てる中で最も柔らかい状態を実現するように, Hip-Joint, Thigh-JointのPD制御ゲイン($K_p$, $K_d$)は(50, 2), Calf-Jointは(30, 0.2)とした.
  観測はlegged\_gym\cite{rudin2022leggedgym}のデフォルト設定に従い, それらのパラメータは\url{https://github.com/haraduka/mevius}を参照されたい.

  MEVIUSにおける強化学習で特に重要だった点は通信遅延である.
  構成のシンプルさを求め, 1チャンネルのCAN-USBインターフェースで12個のサーボモータを動かしている.
  そのため, 全体の制御周期は150Hz, 強化学習の実行周期は50Hzでスレッドとして非同期に回っており, 直接マイコンやFPGAから動かすロボットに比べると通信遅延が大きい.
  通信遅延を考慮せずに強化学習を実行し実機に導入すると, 激しい振動によりロボットの脚が壊れてしまう.
  \figref{figure:exp-graph}がその際の右前の脚の指令関節角度$\bm{\theta}^{ref}_{\{hip, thigh, calf\}}$と実際の関節角度$\bm{\theta}^{cur}_{\{hip, thigh, calf\}}$の様子である.
  シミュレータ上ではほぼピッタリ追従する指令値と実際の関節角度が, 実機では時間方向に大きくシフトしていることがわかる.
  そこで, 本研究では強化学習におけるActionの反映を数ステップずらすことで, これに対処した.
  予備実験からMEVIUSの通信遅延は約0.02$\sim$0.06 secであることが分かったため, ランダムに1$\sim$3フレームだけアクションを遅らせた状態で学習を行った.
}%

\begin{figure*}[t]
  \centering
  \includegraphics[width=1.9\columnwidth]{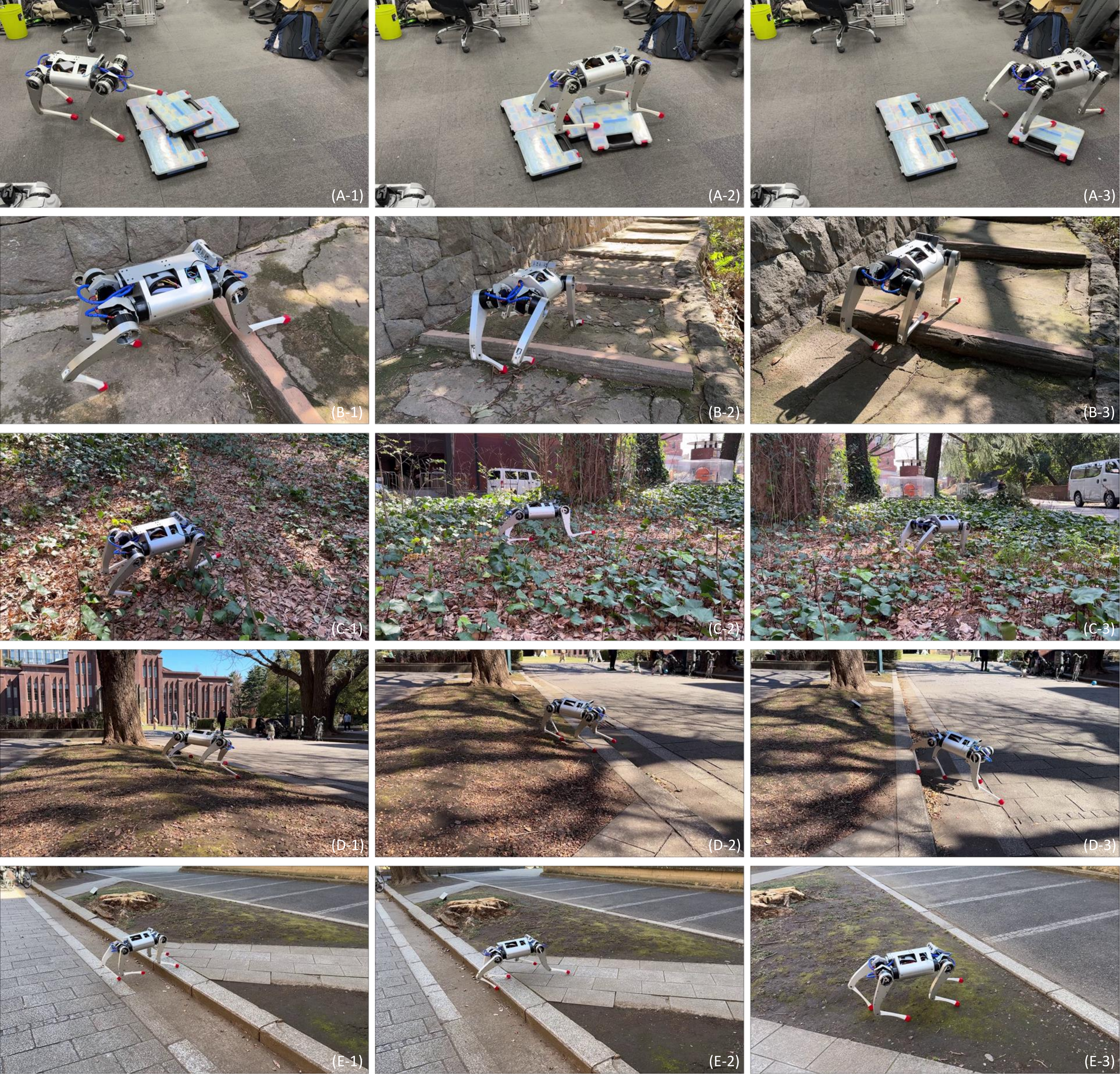}
  \caption{Traversing experiments in various environments: (A) Changing indoor terrain, (B) Gentle upward incline with small steps, (C) Grassy area, and (D, E) Steps with a mixture of soil and cobblestones.}
  \label{figure:exp-act}
\end{figure*}

\begin{figure}[t]
  \centering
  \includegraphics[width=0.95\columnwidth]{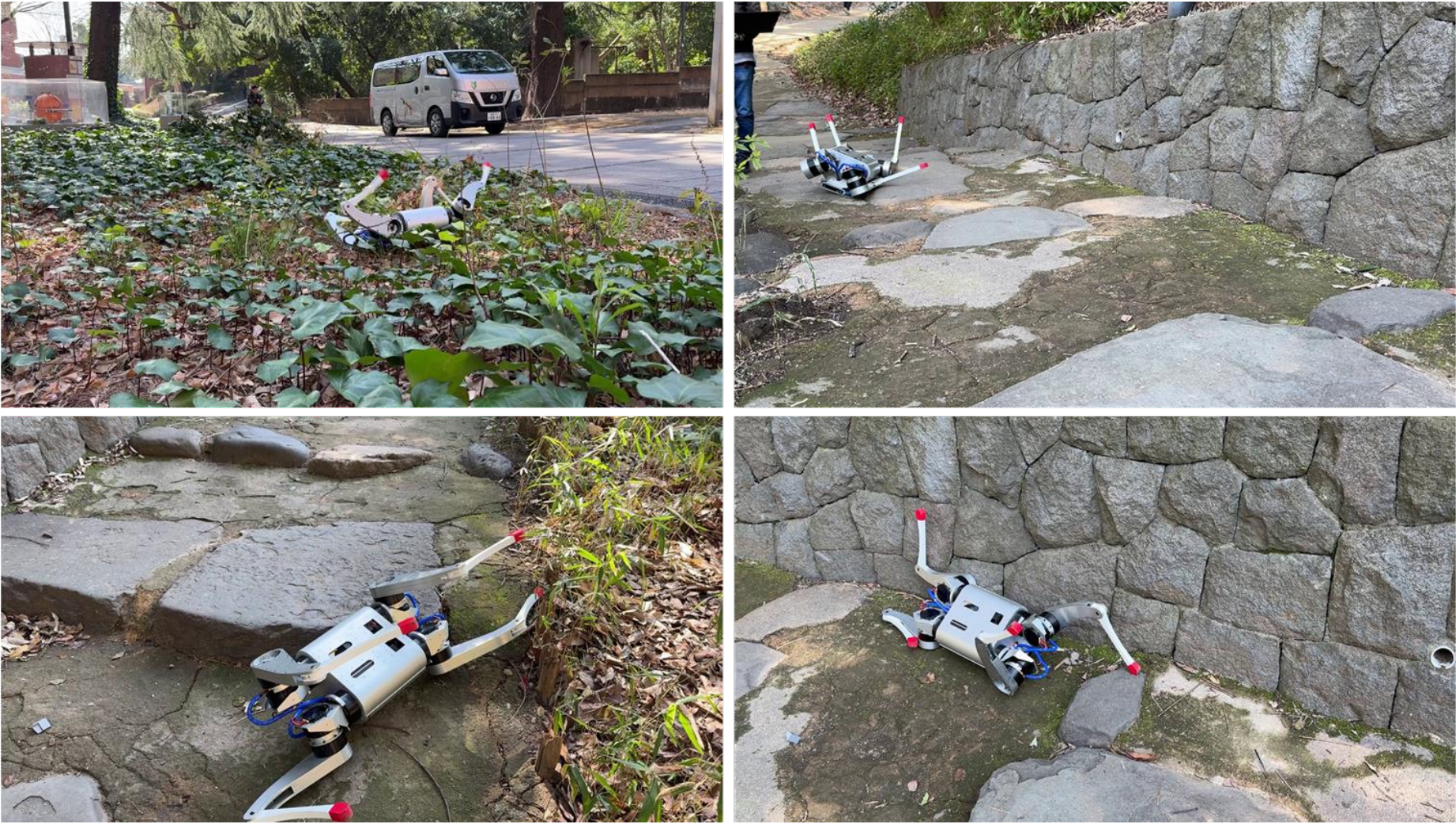}
  \caption{MEVIUS fell due to high speed or high steps. Even in such a situation, none of the body or leg components were damaged.}
  \label{figure:failure}
\end{figure}

\section{Experiments} \label{sec:experiment}

\switchlanguage%
{%
  The results of deploying reinforcement learning in real-world environments are shown in \figref{figure:exp-act}.
  (A) is a changing indoor terrain, (B) is a gentle upward incline with steps, (C) is a grassy area, and (D) and (E) are steps with a mixture of soil and cobblestones. In all environments, the robot successfully navigated while maintaining balance, demonstrating its capability to handle diverse ground surfaces, inclines, and steps.
  In more challenging experiments involving steeper steps or higher speeds, there were instances of the robot stumbling (\figref{figure:failure}).
  Even in these cases, the damaged components were limited to 3D-printed PLA parts used for attaching the Intel Realsense T265, affirming the effectiveness of a robust body structure made of metal and POTICON.
}%
{%
  強化学習の結果を実環境にデプロイした際の結果を\figref{figure:exp-act}に示す.
  (A)は室内における変化する足場, (B)は緩やかな上り坂の参道, (C)は草むら, (D)と(E)は土と石畳が入り交じる段差における踏破実験である.
  どの環境についてもロボットはバランスを取りながら踏破に成功しており, 多様な地面や傾斜, 段差に対応できることが分かった.

  一方で, より険しい段差や速いスピードでの実験では, 一部転んでしまうケースがあった.
  その際も, 壊れた部品はRealsense T265を取り付けるPLAの3Dプリンタ製造部品のみであり, 金属とPOTICONによる頑健な身体構成の有効性を外部環境実験を通して確認することができた.
}%

\section{CONCLUSION} \label{sec:conclusion}
\switchlanguage%
{%
  In this study, we have developed a quadruped robot, MEVIUS, that can be constructed entirely through e-commerce using metal machining, sheet metal welding, and off-the-shelf circuit components.
  We considered the minimal configuration for constructing a quadruped robot and implemented a simple and understandable design, along with the construction of a control system.
  In particular, the torso and some components are integrated through sheet metal welding, allowing for a significant reduction in the number of parts while keeping costs lower compared to machining.
  Furthermore, we applied reinforcement learning considering communication delays to MEVIUS, successfully enabling it to traverse various uneven terrains in real-world environments.
  By openly sharing our design as open source, we anticipate that individual researchers can incorporate new customization and mechanisms, leading to the creation of more diverse quadruped robots that are usable in the real world.
  Additionally, we are currently developing a humanoid robot with a similar structure and plan to use it in future robot education activities through experiments in real-world environments.

}%
{%
  本研究では, 金属切削と板金溶接, off-the-shelfな回路部品のみを用いて, e-commerceを通して完結して構築可能な四脚ロボットMEVIUSを開発した.
  四脚ロボットを構成するうえでの最小構成を考え, 簡易でわかりやすい設計, 制御システムの構築を行った.
  特に, 胴体や一部の部品は板金溶接により一体部品となっており, 切削に比べて価格を抑えつつ大きく部品点数を減らすことができる.
  また, 本研究で開発したMEVIUSに通信遅延を考慮した強化学習を適用し, 実環境の様々な不整地を踏破することに成功した.
  設計や制御をオープンソースで公開することで, 研究者個人が新しい機構や制御を取り入れ, より多様で実世界適用可能な四脚ロボットが作られていくことが期待される.
  また, 今後実環境での実験を通したロボット教育活動にも役立てていく予定である.
}%

{
  \bibliographystyle{IEEEtran}
  \bibliography{main}
}

\end{document}